\documentclass[conference]{IEEEtran}
\usepackage{cite}
\usepackage{amsmath,amssymb,amsfonts}
\usepackage{algorithmic}
\usepackage{graphicx}
\usepackage{textcomp}
\usepackage{xcolor}
\usepackage{acronym}
\usepackage{multirow}
\usepackage[ruled]{algorithm2e}
\usepackage{subcaption}
\usepackage{url}
\def\BibTeX{{\rm B\kern-.05em{\sc i\kern-.025em b}\kern-.08em
    T\kern-.1667em\lower.7ex\hbox{E}\kern-.125emX}}

\def\Pos{\mathbf{P}} 
\def\Neg{\mathbf{N}} 
\def\pos{\mathbf{p}}
\def\neg{\mathbf{n}}
\def\noPos{P}
\def\noNeg{N}

\def\Refs{\mathbf{R}}
\def\refs{\mathbf{r}}
\def\noRefs{R}
\def\Test{\mathbf{Y}}
\def\test{\mathbf{y}}
\def\noTest{Y}

\def\one{\mathbf{1}}
\def\Identity{\mathbf{I}}
\def\bkernelmatrix{\tilde{\mathbf{K}}}
\def\cbkernelmatrix{\tilde{\mathcal{K}}}
\def\rspace{\mathbb{R}}
\def\dim{\mathcal{D}}
\def\kernelfunction{\tilde{\kappa}}
\def\kernelmatrix{\mathbf{K}}

\DeclareMathOperator{\mean}{mean}
\DeclareMathOperator{\std}{std}

\begin{document}

\title{Generalized Reference Kernel With Negative Samples For Support Vector One-class Classification \\
}

\author{\IEEEauthorblockN{Jenni Raitoharju}
\IEEEauthorblockA{\textit{Faculty of Information Technology},
\textit{University of Jyväskylä},
Jyväskylä, Finland,
0000-0003-4631-9298}
}

\newacro{CNN}[CNN]{convolutional neural network}
\newacro{ESVDD}[ESVDD]{Ellipsoidal Support Vector Data Description}
\newacro{GRK}[GRK]{Generalized Reference Kernel}
\newacro{GRKN}[GRKneg]{Generalized Reference Kernel with Negative Samples}
\newacro{KDA}[KDA]{Kernel Discriminant Analysis}
\newacro{KPCA}[KPCA]{Kernel Principal Component Analysis}
\newacro{LDA}[LDA]{Linear Discriminant Analysis}
\newacro{OCSVM}[OC-SVM]{One-class Support Vector Machine}
\newacro{NPT}[NPT]{Non-linear Projection Trick}
\newacro{PCA}[PCA]{Principal Component Analysis}
\newacro{RBF}[RBF]{Radial Basis Function}
\newacro{SVDD}[SVDD]{Support Vector Data Description}
\newacro{SVM}[SVM]{Support Vector Machine}

\maketitle
\begin{abstract}
This paper focuses on small-scale one-class classification with some negative samples available. We propose \ac{GRKN} for \ac{OCSVM}. We study different ways to select/generate the reference vectors and recommend an approach for the problem at hand. It is worth noting that the proposed method does not use any labels in the model optimization but uses the original \ac{OCSVM} implementation. Only the kernel used in the process is improved using the negative data.  We compare our method with the standard \ac{OCSVM} and with the binary \ac{SVM} using different amounts of negative samples. Our approach consistently outperforms the standard \ac{OCSVM} using \acl{RBF} kernel. When there are plenty of negative samples, the binary \ac{SVM} outperforms the one-class approaches as expected, but we show that for the lowest numbers of negative samples the proposed approach clearly outperforms the binary \ac{SVM}. 
\end{abstract}
\begin{IEEEkeywords}
One-class Support Vector Machine, negative samples, Generalized Reference Kernel
\end{IEEEkeywords}
\section{Introduction}
\label{sec:intro}

\acresetall

One-class classification techniques are used in situations where samples only from a single (positive) class are available, but it is necessary to create a model for recognizing outliers (negative samples) \cite{seliya2021literature}. In recent years, deep learning-based techniques, e.g., \cite{ruff2018deep}, have been proposed, but also the traditional techniques, such as \ac{OCSVM} \cite{scholkopf1999support} and \ac{SVDD} \cite{tax2004support}, are still commonly used in small-scale problems \cite{zhao2020improved}, \cite{pang2022hybrid} and new variants and extensions \cite{sohrab2023graph, xiao2024privileged} are being proposed.

If information about outliers is available either via expert knowledge or via existence of some negative samples, it should be possible to produce better one-class classification models. This scenario has been extensively considered in large-scale deep learning-based outlier detection \cite{hendrycks2018deep, goyal2020drocc}, but it has received surprisingly little attention in connection to traditional one-class classification techniques. An explanation for this lack of attention could be that these techniques typically perform poorly with negative data. Therefore, when any negative data are available, either binary classifiers or their variants for imbalanced classification, e.g., \cite{tang2008svms}, are used instead.

\ac{SVDD} variant considering negative samples was proposed already along the original \ac{SVDD}\cite{tax2004support}, but the authors concluded that using negative samples in the model optimization often leads to \emph{worse} results than using only positive samples.
More recent \ac{SVDD} variants using negative samples \cite{nguyen2015repulsive, zheng2016smoothly} have been able to boost the performance of the standard \ac{SVDD}, but not to the level of binary \ac{SVM}. In \cite{wang2015solving}, \ac{SVM} and \ac{SVDD} were combined to tackle the problem of one-class classification with negative samples. Extensions of \ac{OCSVM} to consider negative samples have been also proposed  \cite{suvorov2013osa, nguyen2014kernel}, but the implementations are not publicly available. A larger number of one-class classification studies, such as \cite{ding2016ensemble, sohrab2023graph}, have used negative samples for model validation (hyperparameter selection), while they are not used in the core model optimization. In such works, it is typically not evaluated how critical the availability of representative negative samples is for the model performance. Finally, some works have compared one-class and binary classification methods in different settings \cite{mei2015novel, oosterlinck2020one}.

In this paper, we propose a new method for small-scale one-class classification with negative samples. Our method is based on \ac{OCSVM} and \ac{GRK} \cite{raitoharju2022referencekernel_IJCNN}. We propose a novel approach for selecting \ac{GRK} reference vectors based on negative training samples in a manner that can significantly and consistently improve the method performance.

\section{Generalized Reference Kernel with Negative Samples for \acs{OCSVM} }
\label{sec:grk}

\acf{GRK} \cite{raitoharju2022referencekernel_IJCNN} has methodological similarities with approximate kernel approaches, such as random sampling methods \cite{drineas2005nystrom, gittens2016revisiting}, random projection methods \cite{ailon2009fast}, and random Fourier features \cite{rahimi2007random, liao2020random}. However, the main idea of \ac{GRK} is significantly different. The approximate kernel approaches are designed for large datasets, where the original kernel approaches are computationally too expensive. They aim at finding a computationally lighter way to approximate the original kernel matrix or function. \ac{GRK}, on the other hand, was proposed for one-class classification tasks where the amount of data is low. Instead of trying to approximate the original kernel, \ac{GRK} aims at providing a better kernel using implicit data augmentation via reference vectors used in the kernel computation. As the reference vectors do not need to be labeled, the problems of data augmentation in the absence of negative data can be avoided.

In \cite{raitoharju2022referencekernel_IJCNN}, it was shown that selecting the reference vectors in \ac{GRK} as (positive) training samples augmented with random vectors generated from the training data distribution (assuming standard normal distribution for the standardized training set) could improve the results for \ac{SVDD}. For \ac{OCSVM}, data augmentation by sampling reference vectors from the distribution of the positive data was not equally successful, but the initial results suggested that selecting negative samples as reference vectors could improve the results more. However, the potential of using \ac{GRK} with negative samples was not further studied.

In this paper, we propose \ac{GRKN}. The proposed kernel can be used instead of the regular, usually \ac{RBF}, kernel in the standard kernel \ac{OCSVM}, when some negative training samples are available. Our hypothesis is that if we have some negative samples, we can use them to generate more samples from the approximated negative class distribution and improve the \ac{OCSVM} performance by setting them as reference vectors for \ac{GRKN}. Indeed, our experiments presented in Section \ref{sec:experiments} show that by setting the reference vectors as negative samples augmented by vectors sampled from the approximated negative distribution, we can significantly boost the \ac{OCSVM} performance. 

The full algorithm for computing the \ac{GRKN} matrix for \ac{OCSVM} with the proposed reference vector selection approach is provided in Algorithm 1. As in \cite{raitoharju2022referencekernel_IJCNN}, we use a tilde to denote the base kernel and related terms used in the process. Derivations for the basic \ac{GRK} operations, such as centering of the matrices, can be found in \cite{raitoharju2022referencekernel_IJCNN}. It is worth emphasizing that the final \ac{GRKN} kernel size is equivalent to the \ac{RBF} kernel size, i.e., $P\times P$, where $P$ is the number of positive training samples, irrespective of the number of reference vectors used in computing the kernel. The negative and generated samples are not used directly in the \ac{OCSVM} model optimization, but only to form a more discriminative representation for the positive training samples. 

After forming the \ac{GRKN} kernel matrix, the original kernel \ac{OCSVM} implementation can be used for one-class classification.  As it is a well-known method, we skip its details here. 

\begin{table}[tbp]
\caption{\ac{GRK}/\acs{GRKN} variants used in the experiments \vspace{-10pt}}
\label{tab:variants}
\begin{center}
\begin{tabular}{|c|l|}
\hline
\bf{Variant}& \bf{Reference vectors} \\
\hline

1 & $P$ positive training samples \\
2 & $P$ positive + $N$ negative training samples \\
3 & $N$ negative training samples \\
4 & $P$ positive and $N$ negative generated samples \\
5 & $P+N$ negative generated samples \\
6 & $P$ non-positive and $N$ negative generated samples \\
7 & $P$ negative generated samples and \\ & \hspace{5px} $N$ negative training samples (Proposed) \\
8 & $2P$ negative generated samples and \\
& \hspace{5px} $N$ negative training samples \\
9 & $P/2$ negative generated samples and \\
& \hspace{5px} $N$ negative training samples \\
\hline
\end{tabular}
\end{center}
\end{table}

\begin{algorithm}[p]
\label{alg:grk}
\SetAlgoLined
\caption{\acf{GRKN} for \acs{OCSVM}}
\SetAlgoLined
\vspace{5pt}
\textbf{Training}\\
\hrulefill \\
\vspace{5pt}
\SetKwInOut{Input}{Input}
\SetKwInOut{Output}{Output}
\Input{$\Pos = [\pos_1, ..., \pos_{\noPos}] \in \rspace^{\dim \times \noPos}$, \%\emph{Pos. train data}\\
$\Neg = [\neg_1, ..., \neg_{\noNeg}] \in \rspace^{\dim \times \noNeg}$, \%\emph{Neg. train data}\\

$\kernelfunction(\cdot,\cdot)$, \%\emph{Base kernel function}}

\vspace{2mm}
\Output{$\kernelmatrix_{\Pos \Pos} \in \rspace^{\noPos \times \noPos}$, \%\emph{\ac{GRK} matrix}}
 \vspace{3mm}

\%\emph{Compute mean and std of negative samples}\\
$\mu_{neg} = \mean( \neg_1, \dots, \neg_{\noNeg}),$\\  
$\sigma_{neg} = \std( \neg_1, \dots, \neg_{\noNeg})$\\
 \vspace{3mm}
 
\%\emph{Sample $\noPos$ random vectors}\\
$\mathbf{M} \in \rspace^{\dim \times \noPos} \sim \mathcal{N}(\mu_{neg},\sigma_{neg}^2)$
 \vspace{3mm}
 
\%\emph{Collect reference vectors to $\Refs \in \rspace^{\dim \times (\noNeg+\noPos)=\dim \times \noRefs}$}\\
$\Refs = [\Neg; \mathbf{M}]$
 \vspace{3mm}

\%\emph{Compute uncentered $\bkernelmatrix_{\Refs \Refs}\in \rspace^{\noRefs \times \noRefs}$ and center it}\\
$[\bkernelmatrix_{\Refs \Refs}]_{ij}  = \kernelfunction(\refs_i,\refs_j)$ 
$\cbkernelmatrix_{\Refs \Refs} = (\Identity - \frac{1}{\noRefs}\one_{\noRefs}\one_{\noRefs}^T)\bkernelmatrix_{\Refs \Refs}(\Identity - \frac{1}{\noRefs}\one_{\noRefs}\one_{\noRefs}^T)$
 \vspace{3mm}

\% \emph{Calculate the eigendecomposition $\cbkernelmatrix_{\Refs \Refs}$}\\
 $\cbkernelmatrix_{\Refs \Refs} = \mathbf{U} \boldsymbol{\Lambda} \mathbf{U}^{-1} = \mathbf{U} \boldsymbol{\Lambda} \mathbf{U}^T$
\vspace{3mm}

\%\emph{Compute the pseudoinverse of $\cbkernelmatrix_{\Refs \Refs}$} using the $r$ non-zero eigenvalues\\
 $\cbkernelmatrix_{\Refs \Refs}^{+} = \mathbf{U}_r \boldsymbol{\Lambda}_r^{-1} \mathbf{U}_r^T,$
\vspace{3mm}

\%\emph{Compute uncentered $\bkernelmatrix_{\Refs \Pos} \in \rspace^{\noRefs \times \noPos}$ and center it }\\
$[\bkernelmatrix_{\Refs  \Pos}]_{ij}  = \kernelfunction(\refs_i,\pos_j)$\\
$\cbkernelmatrix_{\Refs  \Pos} = (\Identity - \frac{1}{\noRefs}\one_{\noRefs}\one_{\noRefs}^T)\left(\bkernelmatrix_{\Refs \Refs} - \bkernelmatrix_{ \Refs \Pos}(\frac{1}{\noRefs}\one_{\noRefs}\one_{\noPos}^T)\right)$
 \vspace{3mm}

\%\emph{Compute the \acs{GRKN} matrix} \\
$\kernelmatrix_{\Pos \Pos} = \cbkernelmatrix_{\Refs \Pos}^T\cbkernelmatrix_{\Refs \Refs}^{+}\cbkernelmatrix_{\Refs \Pos} = \cbkernelmatrix_{\Pos \Refs}\cbkernelmatrix_{\Refs \Refs}^{+}\cbkernelmatrix_{\Refs \Pos}$\\
 \vspace{3mm}
\hrulefill \\
\vspace{5pt}
\textbf{Testing}\\
\hrulefill \\
\vspace{5pt}
\SetKwInOut{Input}{Input}
\SetKwInOut{Output}{Output}
\Input{$\cbkernelmatrix_{\Pos\Refs}, \cbkernelmatrix_{\Refs \Refs}^{+}, \Refs$, \%\emph{Computed in training}\\
$\Test \in \rspace^{\dim \times \noTest}$, \%\emph{Test data} \\
}
\vspace{2mm}
\Output{$\kernelmatrix_{\Pos \Test} \in \rspace^{\noPos \times \noTest}$, \%\emph{\ac{GRK} matrix for test data}}
\vspace{3mm}

\%\emph{Compute uncentered $\bkernelmatrix_{\Refs \Test} \in \rspace^{\noRefs \times \noTest}$ and center it }\\
$[\bkernelmatrix_{\Refs  \Test}]_{ij}  = \kernelfunction(\refs_i,\test_j)$\\
$\cbkernelmatrix_{\Refs  \Test} = (\Identity - \frac{1}{\noRefs}\one_{\noRefs}\one_{\noRefs}^T)\left(\bkernelmatrix_{\Refs \Refs} - \bkernelmatrix_{ \Refs \Test}(\frac{1}{\noRefs}\one_{\noRefs}\one_{\noTest}^T)\right)$
 \vspace{3mm}

\%\emph{Compute the \acs{GRKN} matrix for test data} \\
$\kernelmatrix_{\Pos \Test} = \cbkernelmatrix_{\Pos \Refs}\cbkernelmatrix_{\Refs \Refs}^{+}\cbkernelmatrix_{\Refs \Test}$
 \vspace{3mm}
\end{algorithm}

\section{EXPERIMENTAL SETUP}
\label{sec:setup}

\subsection{\acs{GRKN}/\acs{GRK} Variants and Comparative Methods}

To find the best way to use the negative training samples in \ac{GRKN} kernel computation, we compared different ways of selecting the reference vectors as summarized in Table \ref{tab:variants}. Some of the compared approaches use also positive training samples as in the original \ac{GRK}. Positive generated samples were randomly sampled from a normal distribution with zero mean and unit variance (data have been standardized wrt. positive training data), negative generated samples were randomly sampled from a normal distribution having the same mean and variance as negative samples as in Algorithm 1. Non-positive generated samples in variant 6 were generated first as positive samples and then 0.5 was added to the absolute value of each element to push the values further away from the origin. Here, the idea was to generate samples that are on the outskirts of the positive distribution without using negative samples.  We also compared our approach with the standard kernel \ac{OCSVM} algorithm and with binary \ac{SVM} classifier.

\subsection{Implementation and Computing Environment}

In all our experiments, we used Matlab R2017b and \ac{SVM} implementation from LIBSVM library\footnote{\url{https://www.csie.ntu.edu.tw/~cjlin/libsvm/}}. For binary \ac{SVM}, we used the C-SVC (-s 0) and, for \ac{OCSVM}, the one-class SVM implementation (-s 2). To be able to use our custom kernel and to ensure similar hyperparameter setting in all the \ac{RBF} kernels including those used as base kernels in \ac{GRK}, we manually compute all the kernels (-t 4). Our implementation is available at
\url{https://github.com/JenniRaitoharju/GRKneg}.

\subsection{Datasets and Negative Sample Selection}
\label{ssec:datasets}

For our experiments, we selected publicly available datasets from UCI Machine Learning Repository\footnote{\url{http://archive.ics.uci.edu/ml}}. We created one-class classification tasks by considering each class as the positive class and the other class(es) as the negative class. A summary of the dataset properties is shown in Table~\ref{tab:datasets}. We used 70\% of each class for training and 30\% for testing. This splitting was done 5 times. In all tasks, the data was standardized with respect to the positive training data. The hyperparameters were set using a cross-validation within the training set as further explained in Section \ref{ssec:hyperparams}. 

To study the impact of different amounts of negative data, we selected randomly, but keeping the same selection for all methods, 5, 10, 20, 30 or all the negative samples to be used in the experiments. Thus, we had 14 different one-class classification tasks and, for each, 5 different subtasks with a different number of negative samples. Furthermore, each task and subtask was repeated 5 times over different train-test splittings.   

\begin{table}[tbp]
\caption{Datasets and subtasks used in the experiments \vspace{-10pt}}
\label{tab:datasets}
\begin{center}
\resizebox{0.95\linewidth}{!}{
\begin{tabular}{|c|cccccc|}
\hline
\bf{Dataset}& $C$ & $N_{tot}$ & $\dim$ & Task abr. & Target class & $\noPos$ \\
\hline
\multirow{3}{*}{Iris} & \multirow{3}{*}{3} & \multirow{3}{*}{150} & \multirow{3}{*}{4} & Iris1 & Setosa & 35\\
&&&& Iris2 & Versicolor & 35\\
&&&& Iris3 & Virginica & 35\\
\hline
\multirow{3}{*}{Seeds} & \multirow{3}{*}{3} & \multirow{3}{*}{210} & \multirow{3}{*}{7} & Seed1 & Kama & 49\\
&&&& Seed2 & Rosa & 49\\
&&&& Seed3 & Canadian & 49\\
\hline
\multirow{2}{*}{Ionosphere} & \multirow{2}{*}{2} & \multirow{2}{*}{351} & \multirow{2}{*}{32} & Ion1 & Good & 157 \\
&&&& Ion2 & Bad & 88\\
\hline
\multirow{2}{*}{Sonar} & \multirow{2}{*}{2} & \multirow{2}{*}{208} & \multirow{2}{*}{60} & Son1 & Rock & 67 \\
&&&& Son2 & Mines & 77 \\
\hline
Qualitative  & \multirow{2}{*}{2} & \multirow{2}{*}{250} & \multirow{2}{*}{6} & Bank1 & No bankr. & 100\\
bankruptcy &&&& Bank2 & Bankr. & 74\\
\hline
Somerville  & \multirow{2}{*}{2} & \multirow{2}{*}{143} & \multirow{2}{*}{6} & Happ1 & Unhappy & 46 \\
happiness &&&& Happ2 & Happy & 53\\
\hline
\multicolumn{7}{l}{\vspace{-5pt}}\\
\multicolumn{7}{l}{$C$ - number of classes, $N_{tot}$- total number of samples,}\\
\multicolumn{7}{l}{$D$ - dimensionality,  Task abr. - task abbreviation in other tables}\\
\multicolumn{7}{l}{$\noPos$ - number of positive training samples in task}
\vspace{-15pt}
\end{tabular}}
\end{center}
\end{table}

\subsection{Hyperparameter Selection and Evaluation Metric}
\label{ssec:hyperparams}

We used 5-fold cross-validation within the training set to evaluate the hyperparameters separately for each subtask described in Section \ref{ssec:datasets}. As we consider only scenarios, where at least few negative training samples are available for training, we avoided the common challenge of one-class classification, where the hyperparameter optimization should be done without negative data. As a part of our experimental setup, we also evaluate how much the number of negative samples affects the final performance of the standard \ac{OCSVM}, where the negative samples are used only for hyperparameter selection.

In all methods and variants, we use the \ac{RBF} kernel defined as 
\begin{equation}
\label{eq:rbf}
\kernelfunction(\pos_i,\pos_j) =  \exp  \left( \frac{ -\| \pos_i - \pos_j\|_2^2 }{ 2\sigma^2 } \right),
\end{equation}
where $\sigma$ is a hyperparameter. In \ac{OCSVM} and \ac{SVM}, the \ac{RBF} kernel was used directly as the main kernel, while \ac{GRKN}/\ac{GRK} variants used it as the base kernel for computing the generalized kernel. We set $\sigma$ in \eqref{eq:rbf} to $\sqrt{sd_{aver}}$, where $d_{aver}$ was the average squared distance between the training samples and $s$ was selected from $s = \{10^{-1}, 10^0, 10^1, 10^2, 10^3\}$. The $C$ for C-SVC implementation was selected from $C = \{10^{-3}, 10^{-2}, 10^{-1}, 10^0, 10^1, 10^2$, $ 10^3\}$ and the $\nu$ for \ac{OCSVM} approaches from $\nu = \{0.05, 0.1,$ $ 0.15, 0.2\}$.

As our evaluation metric, we used Geometric Mean (Gmean), because it considers both True Positive Rate (TPR) and True Negative Rate (TNR). Gmean is defined as $ \text{Gmean}=\sqrt {\text{TPR} \times \text{TNR}}.$

\section{Experimental Results}
\label{sec:experiments}

\begin{figure*}[t!]
    \centering
    \begin{subfigure}[t]{0.5\textwidth}
        \centering
        \includegraphics[width=\linewidth, trim={0.8cm 0 1cm 0},clip]{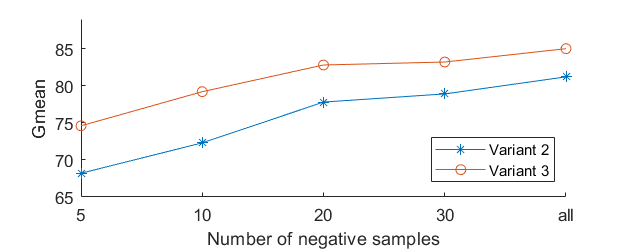}
        \caption{Training samples as reference vectors}
    \end{subfigure}%
    ~ 
    \begin{subfigure}[t]{0.5\textwidth}
        \centering
        \includegraphics[width=\linewidth, trim={0.8cm 0 1cm 0},clip]{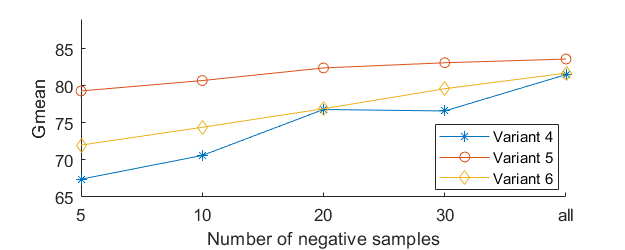}
        \caption{Generated samples as reference vectors}
    \end{subfigure}
    \\
        \centering
    \begin{subfigure}[t]{0.5\textwidth}
        \centering
        \includegraphics[width=\linewidth, trim={0.8cm 0 1cm 0},clip]{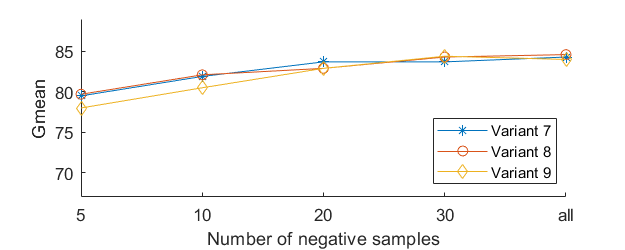}
        \caption{Both training and generated samples as reference vectors}
    \end{subfigure}%
    ~ 
    \begin{subfigure}[t]{0.5\textwidth}
        \centering
        \includegraphics[width=\linewidth, trim={0.8cm 0 1cm 0},clip]{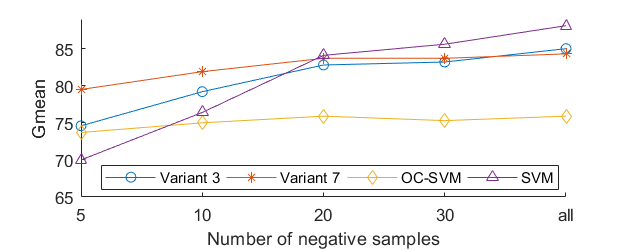}
        \caption{Comparisons with standard methods}
    \end{subfigure}
    \caption{Average Gmean values over all one-class classification tasks }
    \label{fig:cases}
\end{figure*}

Fig.~\ref{fig:cases} shows the average Gmean scores over all the one-class classification tasks with different numbers of negative training samples. Figs.~\ref{fig:cases}a-c show results of \ac{OCSVM} with different \ac{GRK}/\acs{GRKN} variants described in Table~\ref{tab:variants}. Fig.~\ref{fig:cases}-d compares the variants 3 and 7 with the standard \ac{OCSVM} and binary \ac{SVM}. Below, we discuss the results shown in each subfigure.

\emph{Fig.~\ref{fig:cases}-a:  How do different training samples work as reference vectors?}  When only positive training samples were used as reference vectors (variant 1), the performance was so poor (Gmean below 50) that we do not even show the curve in order to keep the same scale for all subfigures. This poor performance is in accordance with the results of \cite{raitoharju2022referencekernel_IJCNN}. The variant 2 using both positive and negative samples works significantly better. However, the variant 3 using only negative samples, even when there are only 5 of them, consistently outperforms the other variants. We can conclude that using positive training samples as reference vectors in \ac{GRK} for \ac{OCSVM} is not recommended.

\emph{Fig.~\ref{fig:cases}-b: If we generate reference vectors instead of using the actual training vectors, how are the results affected? From which distribution should we generate the samples?} As we know the importance of having reference vectors from the negative distribution, variants 4-6 all have $N$ negative generated samples and, in addition, $P$ positive, negative, or non-negative generated samples. We see that having negative generated samples (variant 5) outperforms the other options. Having non-negative generated samples (variant 6) is better than having positive generated samples (variant 4), but not as good as having negative generated samples. On the other hand, approximating the negative distribution with a normal distribution having the same mean and variance must be a really coarse approximation in most cases. We experimented also with some other ways to generate negative samples. For example, we tried adding noise to the negative training samples. However, we could not find anything better than the coarse normal distribution approximation. 

\emph{Fig.~\ref{fig:cases}-c: Is it better to have only generated negative samples or a combination of negative training and generated samples? How does the amount of generated samples affect the results?} Variant 7 is close to variant 5, but $N$ generated negative samples in variant 5 have been replaced with the negative training samples in variant 7. The difference between variants 5 and 7 is small and difficult to see by comparing subfigures b and c, but for all the amounts of negative samples the average Gmean is slightly higher for variant 7. Also the differences between having $P$ (variant 7), $2P$ (variant 8), or $P/2$ (variant 9) negative generated samples is quite small. $P/2$ has a slightly worse performance for the lowest numbers of negative training samples, whereas $P$ and $2P$ generated samples lead to a very similar performance and, therefore, we pick the lower number. As a result of all the comparisons, our proposed approach is the variant 7 that has $P$ negative generated samples and $N$ negative training samples.  

\emph{Fig.~\ref{fig:cases}-d: How are standard \ac{OCSVM} and binary \ac{SVM} affected by the number of negative training samples? Can the proposed approach outperform these approaches?} \ac{OCSVM} is affected by the number of negative training samples only via cross-validation used for hyperparameter selection. The impact is relatively small, but still the average Gmean varies from 73.7 for 5 negative samples to 75.9 for all negative samples. The impact on \ac{SVM} is much larger and, with 5 negative samples its average Gmean is worse than for standard \ac{OCSVM}. The proposed variant 7 clearly outperforms \ac{OCSVM} on all numbers of negative samples and clearly outperforms \ac{SVM} on 5 and 10 negative samples. On 30+ negative samples, the binary \ac{SVM} outperforms all one-class approaches, which is an expected result. Comparison of the variants 3 and 7 shows that having both negative training and generated samples as reference vectors is most beneficial for the lowest numbers of negative training samples, whereas using only negative samples can be a better option when plenty of negative samples are available. As the target use case for the proposed method has only a low number of negative training data, this further confirms that variant 7 is our proposed \acs{GRKN} approach.

\begin{table*}[bt]
\caption{Average Gmean values for \ac{OCSVM}+\ac{GRKN} and comparative methods}
\vspace{-10pt}
\label{tab:numerical}
\begin{center}
\resizebox{1\linewidth}{!}{
\begin{tabular}{|c||c|c|c|c||c|c|c|c||c|c|c|c|}
\hline
& \multicolumn{4}{|c||}{\rule{0pt}{1em} 5 negative samples \rule[-0.5em]{0pt}{1em} } & \multicolumn{4}{|c|}{\rule{0pt}{1em} 10 negative samples \rule[-0.5em]{0pt}{1em} } & \multicolumn{4}{|c|}{\rule{0pt}{1em} 20 negative samples \rule[-0.5em]{0pt}{1em} }\\
\hline
& \ac{SVM} & \ac{OCSVM} & variant 7 & variant 3 & \ac{SVM} & \ac{OCSVM} & variant 7 & variant 3 & \ac{SVM} & \ac{OCSVM} & variant 7 & variant 3 \\
\hline
Iris1 & \bf{100.0$\pm$0.0}& 94.5$\pm$3.2& 97.9$\pm$3.1& 96.6$\pm$3.5 & \bf{100.0$\pm$0.0}& 94.5$\pm$3.2& 97.9$\pm$3.1& 97.6$\pm$2.9 & \bf{100.0$\pm$0.0}& 94.5$\pm$3.2& 99.3$\pm$1.5& 95.9$\pm$2.9\\ 
Iris2 & 80.8$\pm$9.8& 88.9$\pm$7.3& \bf{92.7$\pm$5.9}& 77.4$\pm$8.7 & \bf{92.3$\pm$3.8}& 89.8$\pm$3.5& 88.6$\pm$5.9& 88.5$\pm$3.1 & \bf{96.6$\pm$2.1}& 92.4$\pm$3.5& 90.8$\pm$3.2& 92.2$\pm$3.5\\ 
Iris3 & 93.2$\pm$7.1& 89.1$\pm$1.5& 88.1$\pm$18.3& \bf{95.2$\pm$5.3} & 93.5$\pm$4.3& 86.8$\pm$5.9& 94.0$\pm$7.1& \bf{94.1$\pm$4.8} & \bf{97.0$\pm$2.2}& 89.1$\pm$4.1& 94.1$\pm$4.8& 94.5$\pm$6.0\\ 

Seed1 & 77.3$\pm$10.3& \bf{86.1$\pm$5.6}& 84.7$\pm$8.4& 82.7$\pm$7.3 & 88.0$\pm$5.3& 85.1$\pm$4.5& \bf{91.6$\pm$1.2}& 86.6$\pm$10.2 & 91.3$\pm$3.9& 86.1$\pm$4.9& \bf{93.1$\pm$2.3}& 89.5$\pm$2.4\\ 
Seed2 & 91.0$\pm$4.4& 89.7$\pm$5.9& \bf{93.9$\pm$3.2}& 92.5$\pm$4.0 & \bf{93.5$\pm$2.1}& 90.0$\pm$6.9& 92.7$\pm$3.7& 92.0$\pm$2.1 & \bf{94.7$\pm$1.8}& 91.0$\pm$3.3& 92.2$\pm$2.7& 92.2$\pm$3.3\\
Seed3 & 87.4$\pm$9.6& 91.1$\pm$2.2& \bf{92.8$\pm$1.7}& 90.4$\pm$4.0 & 93.3$\pm$1.7& 91.9$\pm$2.9& \bf{94.0$\pm$2.3}& 92.8$\pm$2.4 & \bf{94.9$\pm$1.0}& 91.6$\pm$3.1& 92.1$\pm$3.2& 93.3$\pm$3.9\\
Ion1 & 48.1$\pm$28.9& 84.4$\pm$8.2& \bf{85.4$\pm$8.4}& 70.5$\pm$9.1 & 72.8$\pm$9.0& \bf{87.2$\pm$4.0}& 83.3$\pm$6.3& 79.1$\pm$5.2 & 81.8$\pm$5.6& 87.9$\pm$2.7& 88.7$\pm$2.2& \bf{89.4$\pm$4.9}\\ 
Ion2 & \bf{65.9$\pm$13.8}& 25.3$\pm$14.4& 65.7$\pm$19.3& 64.0$\pm$11.4 & \bf{75.4$\pm$1.6}& 32.4$\pm$3.1& 72.4$\pm$7.9& 66.3$\pm$11.6 & \bf{84.4$\pm$6.1}& 32.4$\pm$3.1& 79.3$\pm$4.6& 77.4$\pm$5.2\\ 
Son1 & 48.3$\pm$20.8& 52.8$\pm$4.2& \bf{62.0$\pm$10.6}& 61.7$\pm$6.6 & 59.4$\pm$10.4& 54.2$\pm$6.2& \bf{69.6$\pm$2.2}& 55.7$\pm$17.9 & \bf{73.3$\pm$5.6}& 53.2$\pm$4.5& 72.4$\pm$6.4& 73.1$\pm$4.8\\ 
Son2 & 39.0$\pm$24.2& 57.7$\pm$11.0& \bf{62.7$\pm$14.8}& 42.5$\pm$9.0 & 62.2$\pm$5.8& 60.9$\pm$7.0& \bf{68.8$\pm$4.1}& 62.2$\pm$10.3 & 69.4$\pm$4.2& 64.1$\pm$4.8& \bf{71.9$\pm$5.0}& 64.0$\pm$14.9\\ 
Bank1 & 95.3$\pm$3.0& \bf{96.7$\pm$1.6}& 91.6$\pm$4.3& 82.9$\pm$10.4 & 94.3$\pm$3.9& \bf{96.7$\pm$1.6}& 95.9$\pm$3.3& 93.7$\pm$2.6 & 94.7$\pm$2.7& \bf{96.7$\pm$1.6}& 96.0$\pm$1.9& 96.2$\pm$2.2\\ 
Bank2 & 91.6$\pm$9.6& 93.3$\pm$4.9& 93.0$\pm$4.8& \bf{93.7$\pm$1.5} & \bf{96.0$\pm$4.2}& 93.3$\pm$4.5& 93.6$\pm$4.7& 95.9$\pm$2.9 & \bf{98.5$\pm$1.9}& 93.4$\pm$2.7& 95.3$\pm$2.9& 95.0$\pm$3.0\\ 
Happ1 & 40.6$\pm$12.7& 34.2$\pm$6.8& \bf{52.8$\pm$10.6}& 46.5$\pm$12.8 & 27.7$\pm$17.7& 36.2$\pm$5.2& 48.4$\pm$9.0& \bf{52.9$\pm$8.0} & 51.0$\pm$4.3& 38.1$\pm$5.7& \bf{57.0$\pm$5.8}& 56.1$\pm$9.6\\ 
Happ2 & 21.6$\pm$23.1& 48.2$\pm$8.6& \bf{49.9$\pm$11.7}& 48.2$\pm$9.6 & 20.7$\pm$12.7& 51.1$\pm$8.0& \bf{56.4$\pm$3.9}& 51.9$\pm$10.3 & 49.6$\pm$3.7& \bf{51.7$\pm$9.2}& 50.0$\pm$10.5& 49.8$\pm$12.8\\
\hline
Aver. & 70.0$\pm$12.7& 73.7$\pm$6.1& \bf{79.5$\pm$9.0}& 74.6$\pm$7.4 & 76.4$\pm$5.9& 75.0$\pm$4.7& \bf{81.9$\pm$4.6}& 79.2$\pm$6.7 & \bf{84.1$\pm$3.2}& 75.9$\pm$4.0& 83.7$\pm$4.1& 82.8$\pm$5.7\\ 
\hline
\end{tabular}}
\end{center}
\end{table*}

Table~\ref{tab:numerical} shows task-specific numerical results for approaches compared in Fig.~\ref{fig:cases}-d for 5, 10, and 20 negative training samples. While these results show more variation than just the averages shown in Fig.~\ref{fig:cases}, similar conclusions can still be drawn. For 5 negative training samples, \ac{GRKN} variant 7 achieves the best results in most tasks, while with 20 negative training samples binary \ac{SVM} already outperforms it in most tasks.

\section{Conclusions}

This paper proposed a novel approach for small-scale one-class classification with few negative training samples. The proposed approach uses \ac{OCSVM} with \ac{GRK} having negative training and generated samples as reference vectors. The proposed methods clearly outperform standard \ac{OCSVM} and also binary \ac{SVM} for lowest numbers of negative training samples. Wider comparisons of the proposed method with different comparative approaches, such as imbalanced and class-specific classification techniques, remains to be done as future work.



\bibliographystyle{IEEEbib}
\bibliography{refs}

\end{document}